\documentclass[10pt,twocolumn,letterpaper]{article}

\usepackage{times}
\usepackage{epsfig}
\usepackage{graphicx}
\usepackage{amsmath}
\usepackage{amssymb}
\usepackage{amsfonts}
\usepackage{array}
\usepackage{booktabs}
\usepackage{bm}
\usepackage{dcolumn}
\usepackage{float}
\usepackage[utf8]{inputenc}
\usepackage[switch,columnwise]{lineno}
\usepackage{mathtools}
\usepackage{multirow}
\usepackage{mwe}
\usepackage{nicefrac,xfrac}
\usepackage[section]{placeins}
\usepackage{siunitx}
\usepackage{tabularx}
\usepackage{upgreek}
\usepackage{xcolor}
\usepackage{capt-of,etoolbox}
\makeatletter
\apptocmd\@maketitle{{\myfigure{}\par}}{}{}
\makeatother

\usepackage[pagebackref=true,breaklinks=true,letterpaper=true,colorlinks,bookmarks=false]{hyperref}

\newcommand{\keywords}[1]{{\bf \emph{Keywords: #1}}}

\usepackage[capitalize]{cleveref}
\crefname{section}{Sec.}{Secs.}
\Crefname{section}{Section}{Sections}
\Crefname{table}{Table}{Tables}
\crefname{table}{Tab.}{Tabs.}

\newcommand\myfigure{%
\centering
    \begin{tabular}{ccccc}
 \includegraphics[width=0.175\linewidth]{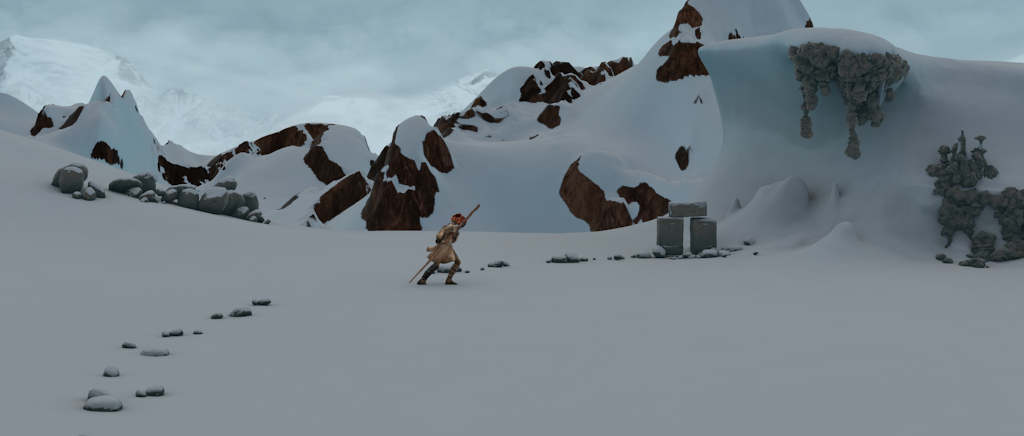}    & 
 \includegraphics[width=0.175\linewidth]{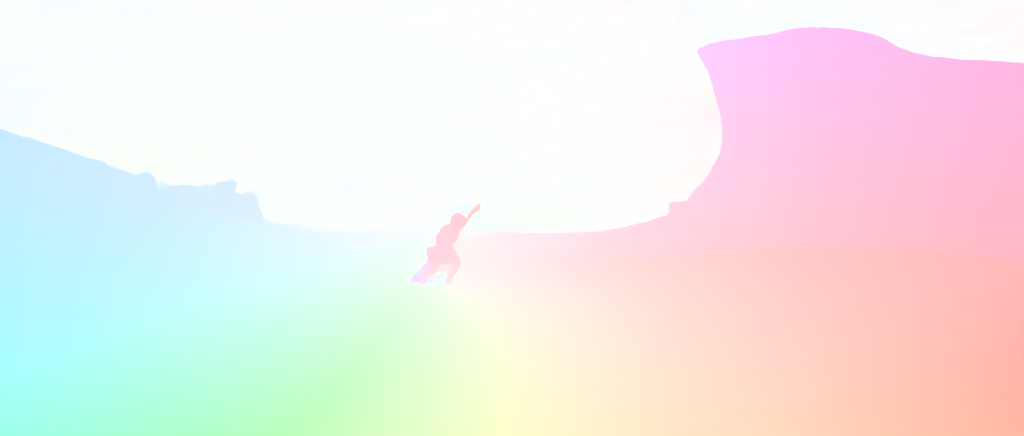} &
 \includegraphics[width=0.175\linewidth]{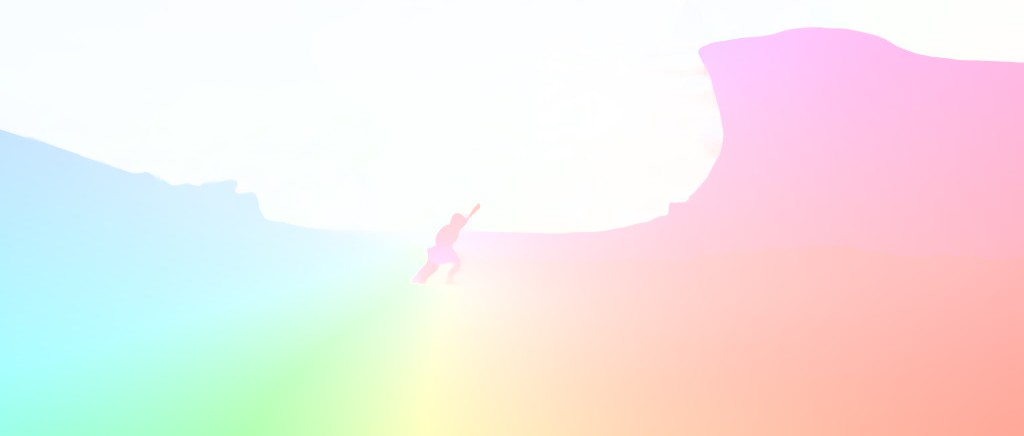} &
 \includegraphics[width=0.175\linewidth]{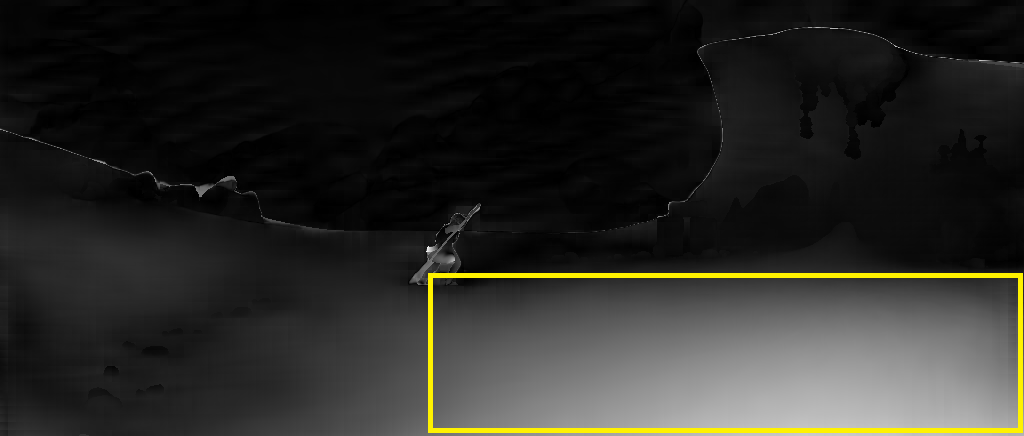}&
 \includegraphics[width=0.175\linewidth]{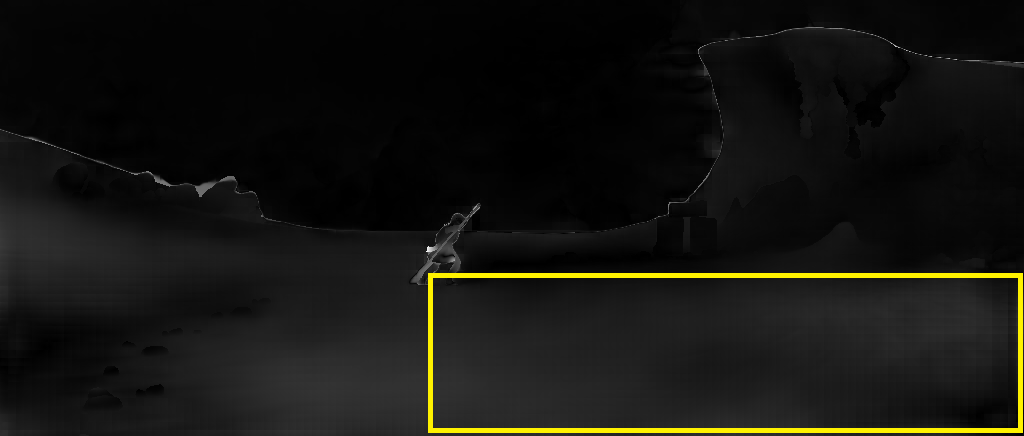}\\
 \includegraphics[width=0.175\linewidth]{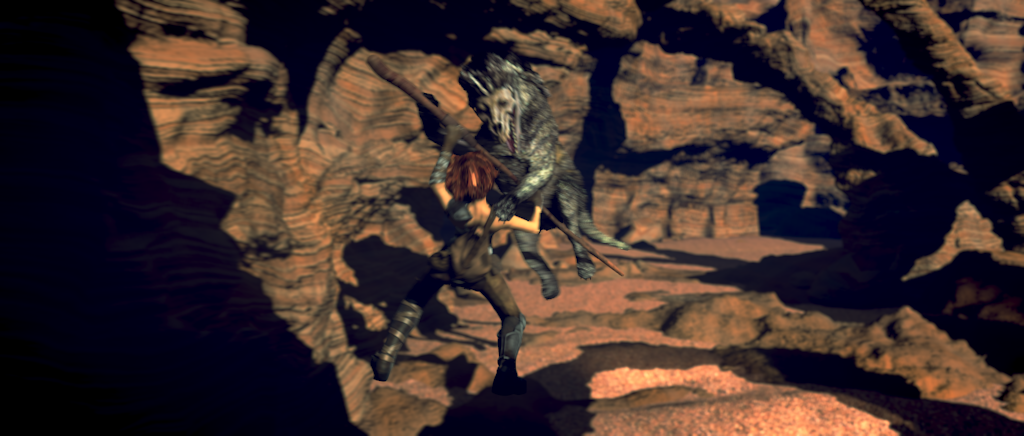} &
 \includegraphics[width=0.175\linewidth]{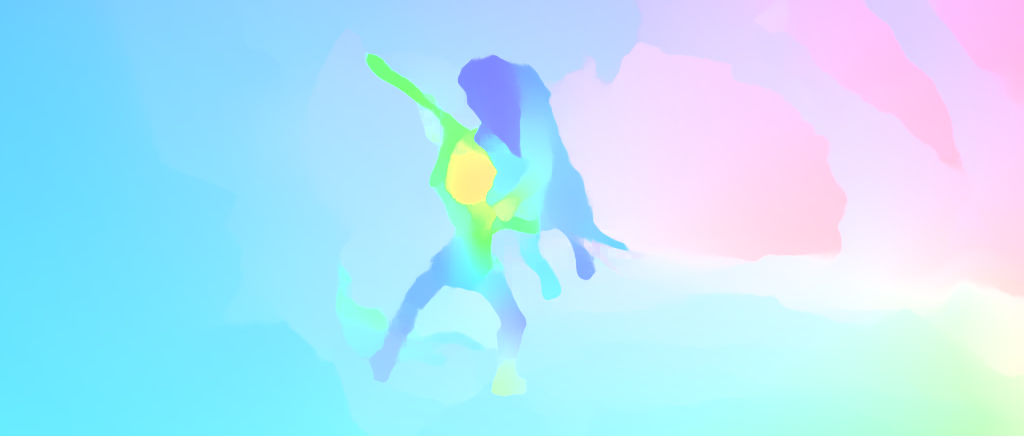}&
 \includegraphics[width=0.175\linewidth]{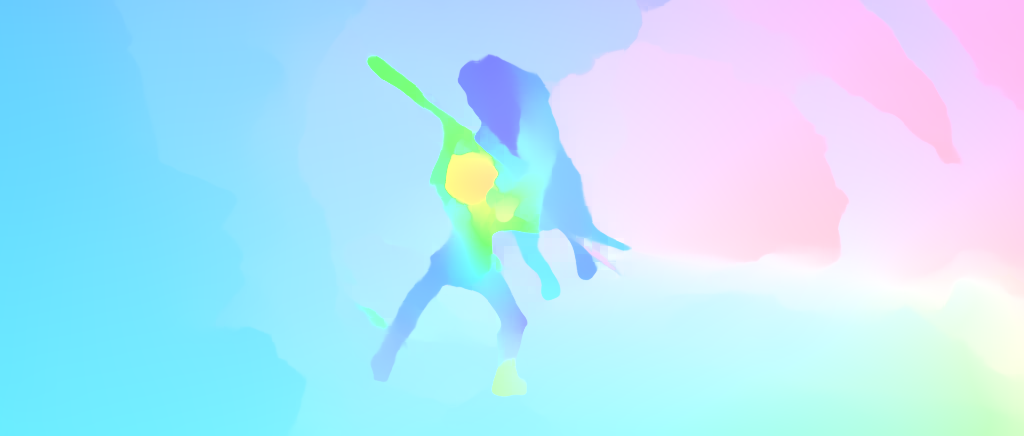}&
 \includegraphics[width=0.175\linewidth]{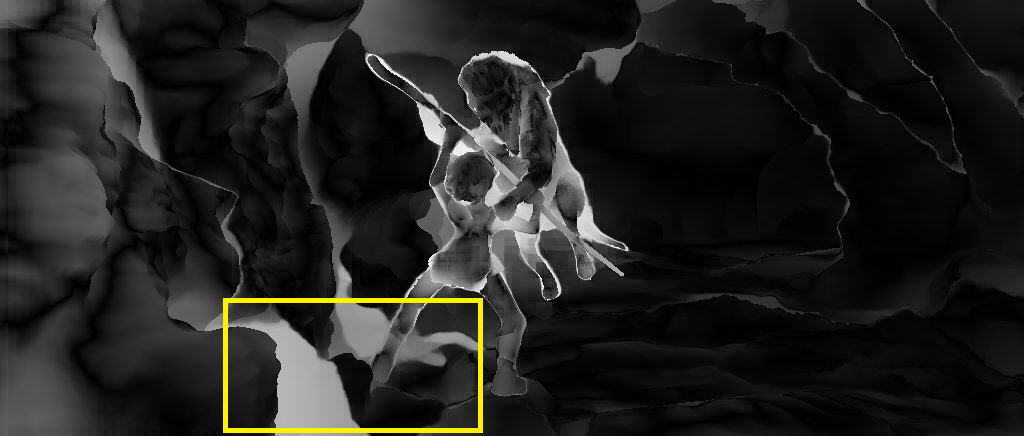}&
 \includegraphics[width=0.175\linewidth]{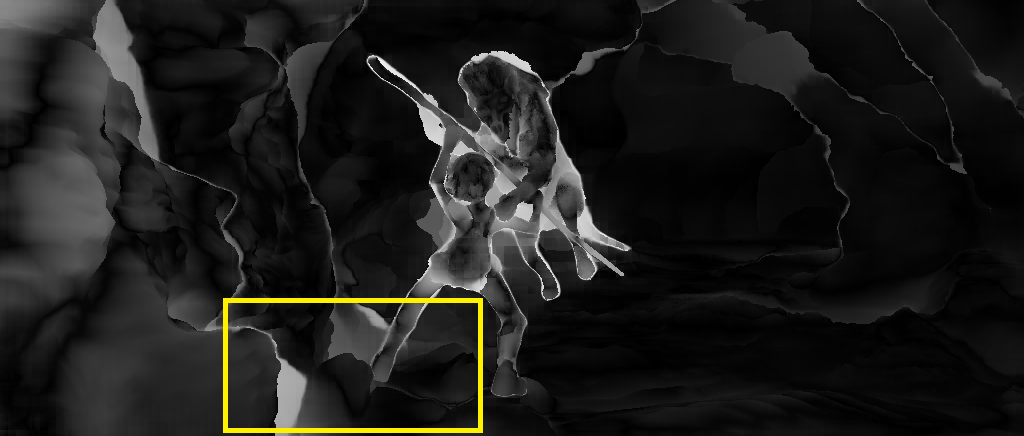}\\
 \includegraphics[width=0.175\linewidth]{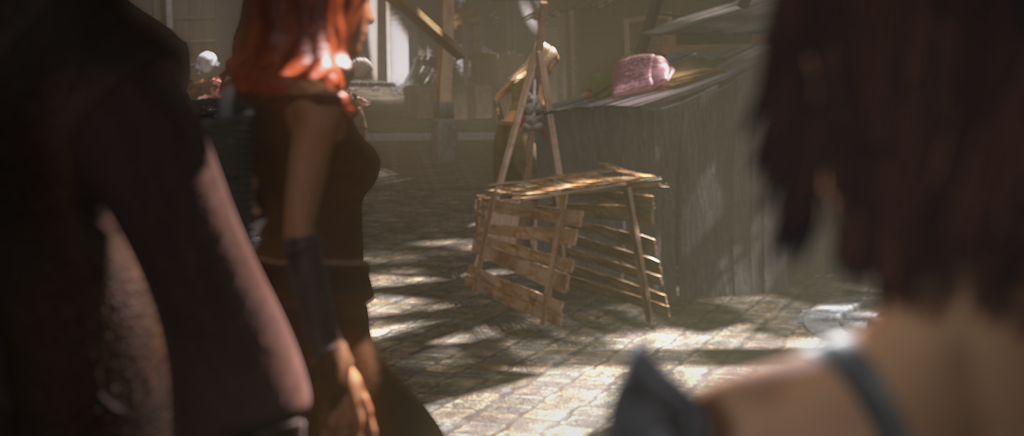} &
 \includegraphics[width=0.175\linewidth]{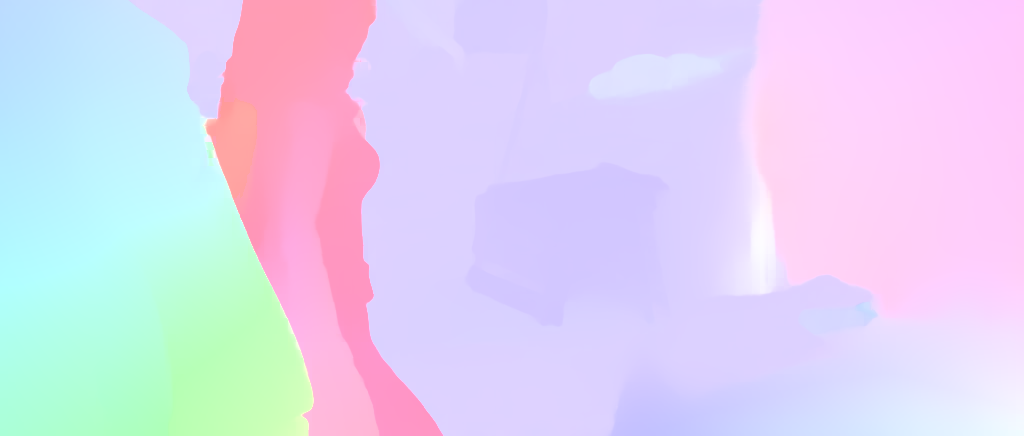} &
 \includegraphics[width=0.175\linewidth]{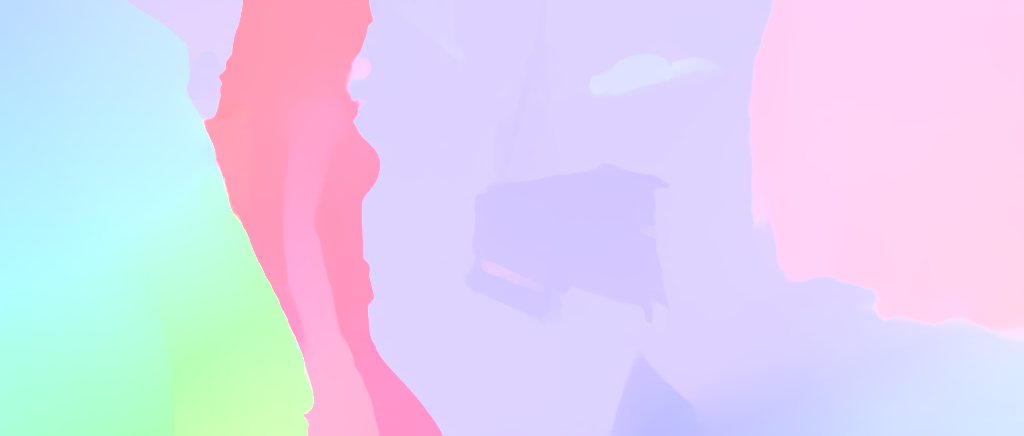}&
 \includegraphics[width=0.175\linewidth]{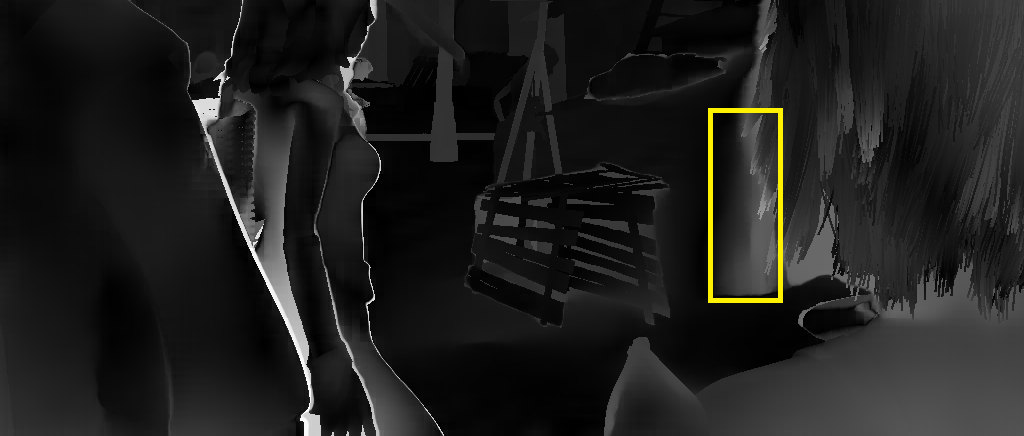}&
 \includegraphics[width=0.175\linewidth]{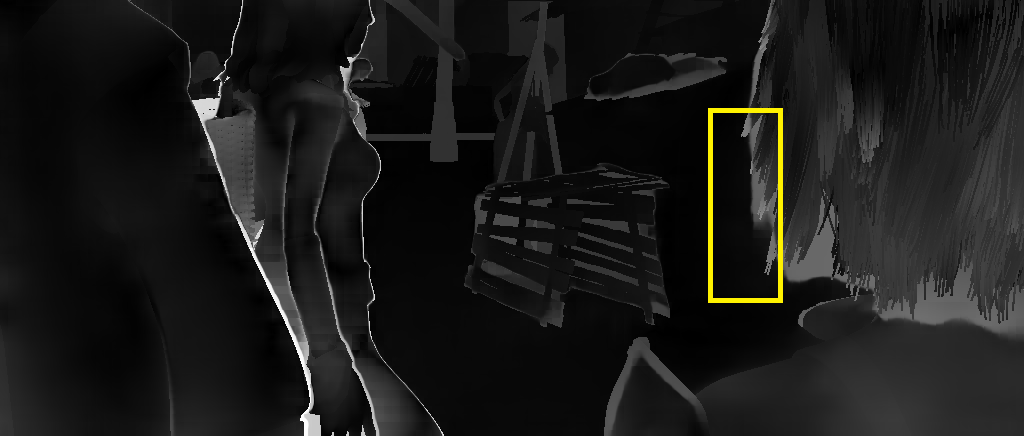}\\
{Reference Frame} & {GMA Flow} & {CGCV Flow} & {GMA Error}& {CGCV Error}
\end{tabular}
\captionof{figure}{We propose to guide the traditional outlier-prone correlation volume by context information. By integrating the proposed Context Guided Correlation Volume with the framework of GMA \cite{Jiang2021GMA}, flow computation accuracy can be successfully improved in challenging cases (see the boxed regions). From top to bottom: Inputs and results for test sequences \textit{Mountain 2}, \textit{Tiger} and \textit{Perturbed Market 3}, the first one is from Sintel (test) Clean dataset, and the last two are from Sintel (test) Final dataset.}
\vspace{3em}
\label{fig:GraphAbs}
}
\begin{document}

\title{CGCV: Context Guided Correlation Volume for Optical Flow Neural Networks}


\author{Jiangpeng Li\qquad Yan Niu\\
\\
Jilin University\\
}

\date{}
\maketitle

\begin{abstract}
   Optical flow, which computes the apparent motion from a pair of video frames, is a critical tool for scene motion estimation. Correlation volume is the central component of optical flow computational neural models. It estimates the pairwise matching costs between cross-frame features, and is then used to decode optical flow. However, traditional correlation volume is frequently noisy, outlier-prone, and sensitive to motion blur. We observe that, although the recent RAFT algorithm also adopts the traditional correlation volume, its additional context encoder provides semantically representative features to the flow decoder, implicitly compensating for the deficiency of the correlation volume. However, the benefits of this context encoder has been barely discussed or exploited. In this paper, we first investigate the functionality of RAFT's context encoder, then propose a new Context Guided Correlation Volume (CGCV) via gating and lifting schemes. CGCV can be universally integrated with RAFT-based flow computation methods for enhanced performance, especially effective in the presence of motion blur, de-focus blur and atmospheric effects. By incorporating the proposed CGCV with previous Global Motion Aggregation (GMA) method, at a minor cost of 0.5\% extra parameters, the rank of GMA is lifted by 23 places on KITTI 2015 Leader Board, and 3 places on Sintel Leader Board. Moreover, at a similar model size, our correlation volume achieves competitive or superior performance to state of the art peer supervised models that employ Transformers or Graph Reasoning, as verified by extensive experiments. 
\end{abstract}


\section{Introduction}
\label{sec:intro}
Given a video sequence, Optical Flow establishes the correspondence between the projection positions of the same scene point onto two consecutive frames. As optical flow computation does not demand information about the camera or the scene, it is probably the only approach to obtain motion clue in applications where only video frames are available, such as graphical rendering \cite{Xu2022TVCG}, augmented reality \cite{Liu2022TVCG}, facial image registration \cite{zhuang2022CVMJ}, video stabilization \cite{yu2020learning, liu2014steadyflow}, action recognition \cite{feichtenhofer2016convolutional, sun2018optical, wang2019hallucinating, diba2019dynamonet, shou2019dmc}. 

Due to its importance, optical flow computation has been intensively investigated since 1980, traditionally modeled by Partial Different Equations \cite{LKpyramidal,compass,dynamically,liu2010sift}, Variation Regularization \cite{horn1981,zach2007duality,werlberger2010,sun2010}, and Statistical Learning \cite{black1993framework,sun2008,roth2007} etc. These models can be viewed as matching pre-described low-level features in a continuous space. Commonly, they focus on explicitly defining discriminative features that are invariant to motion, based on human observations on the distribution nature of video data. The feature invariance and flow smoothness principles are finally modelled by flow interpolation schemes. 

In recent years, research efforts have been largely dedicated to matching high-level features extracted by deep neural models \cite{ilg2017flownet,ranjan2017spy,sun2018pwc,teed2020raft,truong2020glu,zhai2020object,deng2021detail,zhang2021self,hui2020liteflownet3,Jiang2021GMA,sui2022craft,jiang2021cotr,huang2022flowformer, jeong2022imposing, luo2022learning, zhao2022global,xu2022gmflow,zheng2022dip,bai2022deep}. Since Hosni et al introduced the Cost Volume to flow computation \cite{hosni2012fast}, modern parameterized flow computation neural models commonly perform feature matching by a 4D cost volume (or the dual Correlation Volume). which estimates the affinity between cross-frame pairs of features locally or globally. Entries of an ideal correlation volume should be and only be large at true positive correspondence positions. Unfortunately, the correlation volume is generally noisy and outlier-prone, frequently causing matching ambiguity in the presence of motion blur or fast motion. A solution is to down-weight ambiguous matches. To address this issue, in the scope of general multi-image matching, previous works perform $L^2$ normalization to matching features (i.e., the ones that generate the correlation volume\footnote{In this paper, by \textit{matching features}, we mean the features that participate correlation volume construction.}) and the correlation volume \cite{melekhov2019dgc, truong2020glu, rocco2017geometric}; Particularly for optical flow computation, LiteFlowNet3 trains the model to learn a modulator to adjust the correlation vectors and prohibit outliers \cite{ilg2017flownet}. MaskFlowNet trains the model to learn a mask to filter out occluded features \cite{zhao2020maskflownet}. In this paper, we propose a new correlation volume construction strategy, leveraging image brightness and semantic contexts. 

The Recurrent All-Pairs Field Transforms (RAFT) establishes a new flow computation paradigm, in which an additional context branch is designed beside the traditional correlation volume branch \cite{teed2020raft}. RAFT shows that injecting the context features\footnote{In this paper, by \textit{context features}, we mean the features extracted by the context encoder designed by RAFT, unless otherwise specified.} to the flow decoder gains higher accuracy than merely using the correlation volume. In this paper, we analyze the rational behind the benefits of the context branch. We investigate the different behaviors of the Matching Features and Context Features by a thorough empirical study. Based on our observations, we propose guiding the traditional correlation volume by semantic context information. The proposed correlation volume is light-weight, and can be integrated with state-of-the-art flow computation neural networks at negligible extra cost. To show the effectiveness of our Context Guided Cost Volume (CGCV), we take the recent Global Motion Aggregation (GMA \cite{Jiang2021GMA}) flow network as an example baseline. Comprehensive experiments verify that, by incorporating the proposed cost volume with GMA, the flow computation accuracy is significantly improved, especially in the presence of occlusion and motion blur. Compared to other GMA-based peer methods that utilize graph reasoning or Transformer techniques for performance enhancement, our method shows competitive or superior accuracy and simplicity, as verified by comprehensive experiments including the official Sintel and KITTI 2015 Optical Flow Leader Board evaluations.

In summary, our work contributes:
\begin{enumerate}
    \item A new simple, light-weight, plug-and-play approach to constructing high quality correlation volume for flow computation.
    \item An in-depth analysis on the barely noticed contrast between the behaviors of the matching features and context features of the RAFT flow computation framework. 
    \item Evidently improved flow computation performance over the traditional correlation volume counterpart. 
\end{enumerate}

\section{Related Work}
Even within the scope of neural models, the literature of optical flow is broad. This section focuses on the development of Correlation Volume in the framework of supervised learning, which is directly related to ours.

\noindent\textbf{Correlation Volume Construction} Hosni et al. first introduced the Cost Volume to optical flow computation \cite{hosni2012fast}, with each entry representing the cost of matching a pair of cross-frame yet spatially neighbouring pixels. The image correspondence is estimated from the cost volume in a ``winner takes all'' manner (\textit{argmin}). The cost volume thus transforms the \textit{image correspondence} problem into a \textit{feature correspondence} problem. DCFlow extends the cost measurement from the distance in color and structure spaces to deep feature space, and estimates flow from the cost volume via Flow Semi-Global Matching \cite{xu2017accurate}. FlowNet shows that the Cost (more precisely, Correlation) Volume can be constructed and decoded to the flow field in an end-to-end CNN architecture \cite{ilg2017flownet}. Inspired by the pyramidal flow refinement of SPyNet \cite{ranjan2017spy}, LiteFlowNet \cite{hui2018liteflownet, hui2020lightweight} and PWC-Net \cite{sun2018pwc, sun2019models} extend the image pyramid to feature pyramid, constructing a local correlation volume at each level to refine flow hierarchically. Hur-Roth unified the multi-level sub-networks to one encoder-decoder shared across all levels for iterative residual flow refinement \cite{hur2019iterative}. Devon computes multi-scale local correlation volumes by sampling the target frame's full-resolution feature map at various dilated factors, and concatenates them to decode flow \cite{lu2020devon}. Yang-Ramanan proposed a multi-channel correlation volume, each channel for a different feature embedding \cite{yang2019volumetric}. Although the aforementioned correlation volumes are local, they need to be re-computed (generally by vector inner products) each time the flow is refined. Differently, RAFT retrieves the updated local correlation values from a global all-pairs correlation volume, which remains constant once constructed \cite{teed2020raft}. This strategy bypasses computing inner products between features along with flow refinement. State of the art works GMA \cite{Jiang2021GMA}, AGFlow \cite{AGFlow}, Consistency Imposition \cite{jeong2022imposing}, KPA \cite{luo2022learning}, DEQ \cite{bai2022deep} follow the same design of global correlation volume. CRAFT views the correlation volume of GMA-RAFT as a degenerated cross-attention, and extends it to a non-degenerated multi-head one \cite{sui2022craft}. FlowFormer tokenizes the cost volume to cost memory, from which the flow is decoded by Separable Self Attention Transformer \cite{chu2021twins} layers \cite{flowformer}. Global Matching methods GMFlow \cite{xu2022gmflow} and GMFlowNet \cite{zhao2022global} employ Transformers to extract high quality features. Transformers can largely enhance the correlation volume, but take tremendous graphical memory. Xu et al. used 1D attention and correlation in orthogonal directions to achieve 2D correspondence effect, reducing RAFT's 4D correlation volume to a concatenation of two 3D volumes \cite{xu2021high}. Different from previous works, our aim is to address the matching ambiguity suffered by currently most popular cost volume at minimal cost, especially for RAFT-based flow computation methods. 

\noindent\textbf{Filtering Outliers} Although correlation volume acts as the base to most optical flow computation neural networks, it is actually vulnerable to factors such as motion blur, fast motion or large displacement, occlusion or disocclusion. To down-weight ambiguous matches, CNNGM \cite{rocco2017geometric}, DGC-Net \cite{melekhov2019dgc} and GLU-Net \cite{truong2020glu} perform vector $L^{2}$ normalization to feature descriptors and/or channel-wise $L^{2}$ normalization to the correlation volume. MaskFlowNet \cite{zhao2020maskflownet} learns a multiplicative occlusion mask to filter the warped features of the target frame. LiteFlowNet3 \cite{hui2020liteflownet3} modulates each cost vector by an affine transformation, whose coefficients are learned from the concatenation of a confidence map, the cost volume and the reference frame's matching feature map. Seemingly, our method improves the correlation volume by modulation too. However, our formulation is fundamentally different from MaskFlowNet or LiteFlowNet3. First, based on a thorough empirical study on the context and matching features, our gating and lifting tensors are designed to take advantage of the context stream. Thus our modulation is independent of the correlation volume. Oppositely, MaskFlowNet and LiteFlowNet3 learn such tensors from the matching features and their correlations. Moreover, our modulation needs to be computed only once, hence its extra cost on parameters or FLOPs is very limited, whereas MaskFlowNet and LiteFlowNet3 modulate the features or local correlation vectors along with flow refinement. 

\noindent\textbf{Context Features} The terminology ``Context Encoder'' appears in several flow computation publications. However, they mean totally different operations in different publications. In PWC-Net \cite{sun2018pwc}, it is a flow post-processor. In GMFlowNet \cite{zhao2022global}, it generates the features to be matched to form the correlation volume. In RAFT, the context encoder is independent to the correlation volume. It extracts features from the reference frame and inject them to the flow refinement units. Later, GMA applies part of RAFT's context features to aggregate the motion features and shows flow accuracy improvement \cite{Jiang2021GMA}. KPA \cite{luo2022learning} further applies kernel patch attention of the context features to modulate the motion features. This strategy effectively improves the quality of motion features, but as the attention is computed in each flow refinement unit, the required Floating Point Operations (FLOPs) are remarkably increased. Our paper analyzes the context encoder of RAFT by an in-depth empirical study, based on which, we design a new correlation volume guided by semantic contexts. The proposed volume remains constant during flow refinement, refraining from incurring additional parameters or FLOPs.


\section{Approach}
\label{sec:approach}

In this section, we first discuss the weakness of the commonly employed correlation volume, which is constructed from the matching features. We then analyze the barely noticed advantages of the context features in GMA-RAFT. We thoroughly investigate the rational for the different behavior of GMA-RAFT matching and context feature encoders. Based on these studies, we propose a new Context Guided Correlation Volume (CGCV). Finally, we describe the plug-and-play integration of CGCV with state-of-the-art flow computation neural models.

\begin{figure*}[t]
\centering
\begin{tabular}{cccc}
 \multicolumn{2}{c}{\includegraphics[width=0.225\linewidth]{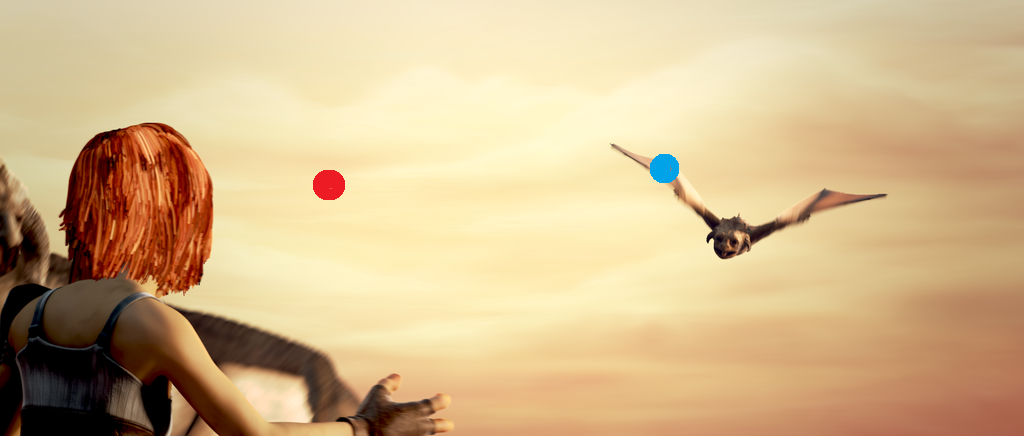}} & 
 \multicolumn{2}{c}{\includegraphics[width=0.225\linewidth]{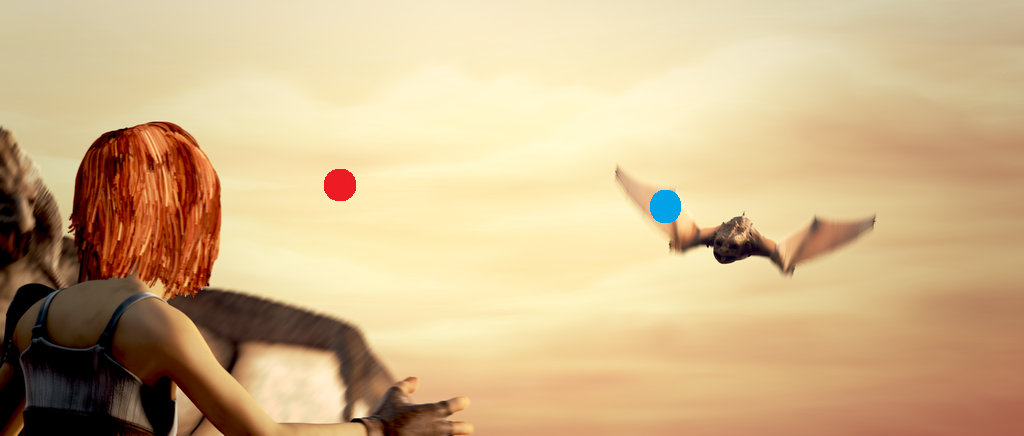}} \\
 \multicolumn{2}{c}{\small{a. Reference frame $\mathbf{I}_{1}$ and two example queries. }} & \multicolumn{2}{c}{\small{b. Target frame $\mathbf{I}_{2}$ and true correspondences of the queries.}}\\
 \multicolumn{2}{r}{\includegraphics[width=0.45\linewidth]{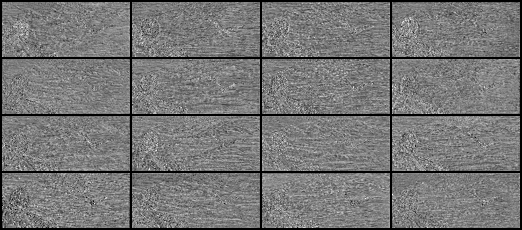}}&
 \multicolumn{2}{l}{\includegraphics[width=0.45\linewidth]{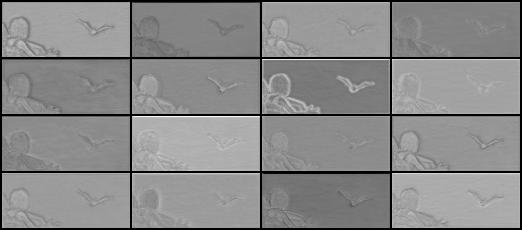}}\\
 \multicolumn{2}{c}{\small{c. The first 16 channels of $g(\mathbf{I}_{1};\theta_{g})$.}} & \multicolumn{2}{c}{\small{d. The first 16 channels for $h(\mathbf{I}_{1};\theta_{h})$.}}\\
 \includegraphics[width=0.225\linewidth]{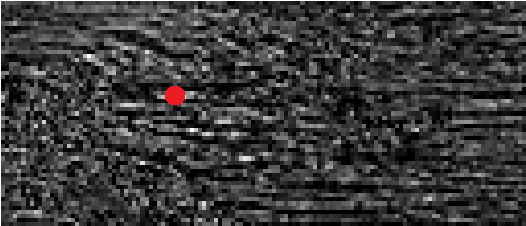}
 &\includegraphics[width=0.225\linewidth]{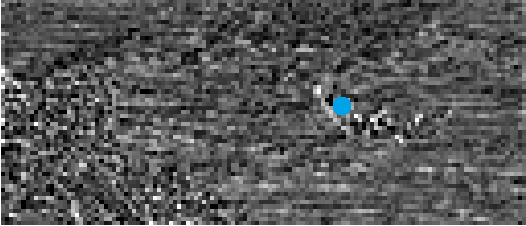}
 &\includegraphics[width=0.225\linewidth]{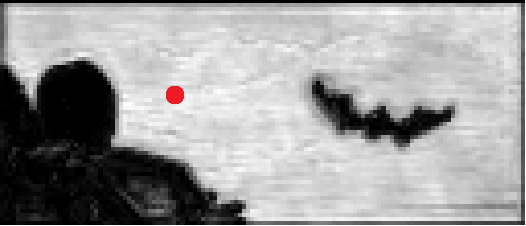} &\includegraphics[width=0.225\linewidth]{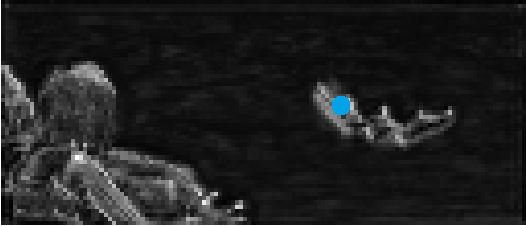}\\
 \multicolumn{2}{c}{\small{e. GMA cost planes. True correspondences are colored.}} & \multicolumn{2}{c}{\small{f. Correlation planes constructed by $h(\mathbf{I}_{1};\theta_{h})$ and $h(\mathbf{I}_{2};\theta_{h})$}}
\end{tabular}
\caption{A comparison between the GMA correlation planes and the correlation planes constructed from the context features, demonstrated by two query points indicated by red and blue dots. The reference and target frame are from sequence \textit{Temple} of Sintel (train) Final benchmark dataset. Dots of the same color on the figures represent a pair of true correspondence points.}
 \label{fig:CostPlanes}
\end{figure*}

\begin{figure*}[h]
\centering
\begin{tabular}{cc}
 \includegraphics[width=0.45\linewidth]{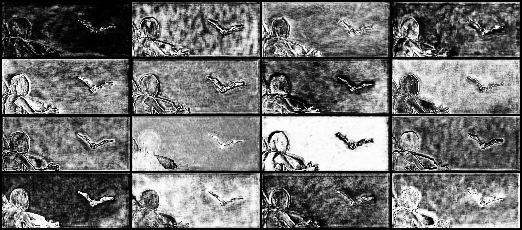}
 &\includegraphics[width=0.45\linewidth]{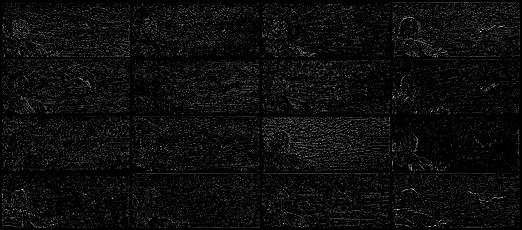}\\
 {a. The first 16 channels of $g^{\text{net}}(\textbf{I}_{1}; \breve{\theta}_{g})$}.
 & {b. The first 16 channels of $g^{\text{inp}}(\textbf{I}_{1}; \breve{\theta}_{g})$}
\end{tabular}
\caption{By sharing matching features $g$ with the flow refinement GRUs, the ``net'' half of the features become contextual. Here the first 16 (out of totally 128) channels of $g^{\text{net}}(\textbf{I}_{1}; \breve{\theta}_{g})$ and $g^{\text{inp}}(\textbf{I}_{1}; \breve{\theta}_{g})$ extracted from the reference frame of sequence \textit{Temple} are presented. See main text for details.}
 \label{fig:InjectingMatch2GRU}
\end{figure*}

\begin{figure*}[t]
\centering
\begin{tabular}{c}
 \includegraphics[width=1.0\linewidth]{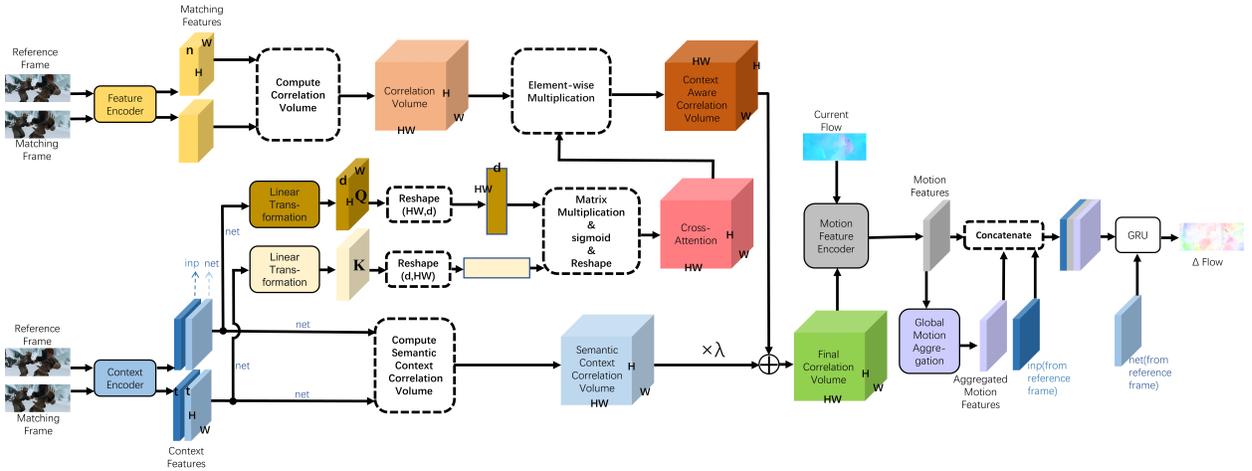}\\
\end{tabular}
\caption{The architecture of our flow computation model, which integrates the proposed CGCV with GMA. }
 \label{fig:CVM}
\end{figure*}
\subsection{Weakness of Traditional Correlation Volume}
Let $\mathbf{I}_{1}, \mathbf{I}_{2}\in\mathbb{R}^{H\times W\times 3}$ be the reference and target frames of a video clip. For each pixel $[i,j]\in \mathbb{N}^2\cap\left\{[1,W]\times [1,H]\right\}$ in $\mathbf{I}_{1}$, we seek for its correspondence point $[i+u,j+v]\in \mathbb{R}^2\cap\left\{[1,W]\times [1,H]\right\}$ in $\mathbf{I}_{2}$, such that $\mathbf{I}_{1}[i,j]$ and $\mathbf{I}_{2}[i+u,k+v]$ are the image projections of the same scene point\footnote{Note that $u$, $v$ vary with $[i,j]$, for conciseness, we abbreviate $u[i,j]$ as $v[i,j]$ as $u$ and $v$.}. To find the optical flow vector $[u,v]\in\mathbb{R}^2$, $\mathbf{I}_{1}[i,j]$ and $\mathbf{I}_{2}[i+u,j+v]$ must share some common feature $\mathbf{F}$ that is representative and invariant to $[u,v]$. That is,  $\mathbf{F}_{1}[i,j]$, as the feature representation for $\mathbf{I}_{1}[i,j]$, equals to $\mathbf{F}_{2}[i+u,j+v]$. Moreover, $\mathbf{F}$ should be an excluding representation, such that for any other displacement $[\Tilde{u},\Tilde{v}]\neq [u,v]$, $\mathbf{F}_{1}[i,j]\neq\mathbf{F}_{2}[i+\Tilde{u},j+\Tilde{v}]$. 

Deep neural methods learn such feature representation by deep networks. For example, RAFT \cite{teed2020raft} designs an encoder $g(~; \theta_{g})$\footnote{To differentiate the training and inference passes of $g$, we let $g(~)$ denote the feed-forward computation with parameters to be trained; $g(~; \theta_{g})$ denote the computation with trained and fixed parameters $\theta_{g}$. Same for the upcoming functions $h(~;\theta_{h})$ and $c(~;\theta_{c})$.} to extract matching features $g(\mathbf{I}_{1}),  g(\mathbf{I}_{2})\in\mathbb{R}^{n\times\frac{H}{8}\times\frac{W}{8}}$, where $n$ is the feature length. The cross-frame all-pairs correlations between $g(\mathbf{I}_{1})$ and $g(\mathbf{I}_{2})$ form a correlation pyramid, whose first level is a correlation volume $\mathbf{C}$ computed by
\begin{equation}
    \mathbf{C}[i,j,k,l]=\frac{1}{\sqrt{n}}
    \langle g(\mathbf{I}_{1})[i,j,1:n], g(\mathbf{I}_{2})[k,l,1:n]\rangle,
\label{eq:CostVolumeLevel1}
\end{equation}
where symbol $\langle\, , \,\rangle$ stands for the inner product of two vectors; $i,k \in [1, \nicefrac{W}{8}]$, $j,l\in[1,\nicefrac{H}{8}]$. Pooing this correlation volume at different scales builds the correlation pyramid, from which, a correlation feature is retrieved for each point $[i,j]$. The correlation feature map is one of the main components to decode the flow field $\hat{\mathbf{U}}, \hat{\mathbf{V}} \in\mathbb{R}^{\frac{H}{8}\times\frac{W}{8}}$. This flow computation framework is widely followed by recent works (e.g., GMA \cite{Jiang2021GMA}, AGFlow \cite{AGFlow}, CRAFT \cite{sui2022craft}). 


Ideally, for each point $\mathbf{x}=[i,j]$, its correlation plane $\mathbf{C}[i,j,1:\frac{W}{8},1:\frac{H}{8}]$ should reach the maximum value at its true correspondence $\mathbf{x}^{\prime}$, and should have a significantly smaller correlation value at any other point $\Tilde{\mathbf{x}}\neq\mathbf{x}^{\prime}$. However, in the presence of occlusion/disocclusion, motion blur, fast motion, etc., the resulted correlation planes may be rather noisy and outlier prone: often exhibiting large correlation values at many false correspondence positions, or small correlation value at the true correspondence, leading to matching ambiguity. Fig.~\ref{fig:CostPlanes}.~a-c and Fig.~\ref{fig:CostPlanes}.~e illustrate such correlation planes by two example points on the reference frame of sequence \textit{Temple} of benchmark dataset Sintel (train) Final, using the trained GMA model. In this example, $g(\textbf{I}_{1}; \theta_{g})$ fails to extract representative matching features from the input frames, and their consequent cross-correlation map is chaotic.

\subsection{What Makes an Encoder Extract Contexts}


To improve the quality of the traditional correlation volume, we pursue a more reliable clue. We notice that, in striking contrast to the matching features, the context features extracted by GMA-RAFT context encoder $h(~;\theta_{h})$ are of much higher quality, and the context cross-frame correlation are significantly sharper and cleaner (see Fig.~\ref{fig:CostPlanes}.~d and Fig.~\ref{fig:CostPlanes}.~f).

An intuitive explanation for their different behaviors is that the context encoder shares features with the flow refinement GRUs, while the matching feature encoder does not. Nevertheless, it should be further noted that different parts of the context features are shared with the GRUs differently in GMA-RAFT. The first 128 channels, termed as ``net'' in the released code of RAFT, initialize the hidden state of the first GRU; The rest, termed as ``inp'',  are input to GRUs. GMA additionally uses the ``inp'' part for the self-attention mechanism to aggregate global motion. For clarity, we denote the two halves by $h^{\text{net}}(\textbf{I}_{1})$ and $h^{\text{inp}}(\textbf{I}_{1})$. Hereafter, we name $g(\textbf{I})$ the ``matching'' features (as they are used for computing matching costs) and $h(\textbf{I})$ the ``context'' features (as in RAFT \cite{teed2020raft}).

To inspect how context information is extracted, we carried out an empirical study by removing the context stream from GMA and using the matching features for replacement. In particular, we split $g(\textbf{I}_{1})$ to two halves
$g^{\text{net}}(\textbf{I}_{1})$ and $g^{\text{inp}}(\textbf{I}_{1})$, which replace $h^{\text{net}}(\textbf{I}_{1})$ and $h^{\text{inp}}(\textbf{I}_{1})$ respectively. We train this experimental model from scratch and denote the learned parameters of $g$ by $\breve{\theta}_{g}$. Fig.~\ref{fig:InjectingMatch2GRU} presents the first 16 feature channels of $g^{\text{net}}(\textbf{I}_{1}; \breve{\theta}_{g})$ and $g^{\text{inp}}(\textbf{I}_{1}; \breve{\theta}_{g})$ extracted from the example image \textit{Temple}. Overall, $g(\textbf{I}_{1}; \breve{\theta}_{g})$ performs significantly better than $g(\textbf{I}_{1}; {\theta}_{g})$ at extracting image details. More particularly, $g^{\text{net}}(\textbf{I}_{1}; \breve{\theta}_{g})$ provides image semantic grouping proposals, although being less smooth than $h^{\text{inp}}(\textbf{I}_{1}; \theta_{h})$; $g^{\text{inp}}(\textbf{I}_{1}; \breve{\theta}_{g})$ detects image brightness variation, with much stronger contrast than $g(\textbf{I}_{1}; {\theta}_{g})$.  
This experiment reveals that, the model training process drives the initial hidden state of the GRU to learn image semantic contexts and drives the GRU inputs to learn image intensity contexts. Even though $g$ operates on both frames, sharing only $g(\mathbf{I}_{1})$ with the GRUs would push the kernels of $g$ to detect different levels of high frequency components of images. 

\subsection{Context Guided Correlation Volume}
Based on the above investigation, we propose guiding the traditional correlation volume by mutual context information, via a cross-attention mechanism. Compared to adding a new modulator to the neural network, our method is advantageous at largely saving extra parameters for feature extraction.

\noindent\textbf{Gating Operation} We extend the context encoder in GMA-RAFT to a Siamese network $c(~)$ to extract features from both the reference and target frames. We partition $c(\mathbf{I})$ to two halves $c^{\text{net}}(\mathbf{I})$ and $c^{\text{inp}}(\mathbf{I})$. $c^{\text{net}}(\mathbf{I}_{1})$ is further used to initialize the hidden state of the flow refinement GRU; $c^{\text{inp}}(\mathbf{I}_{1})$ is used as the input to the GRUs. We define the \textit{Query} matrix $\mathbf{Q}$ as a linear transformation of $c^{\text{net}}(\mathbf{I}_{1})$, and the \textit{Key} matrix $\mathbf{K}$ as a linear transformation of $c^{\text{net}}(\mathbf{I}_{2})$:
\begin{equation}
    \mathbf{Q}  =  \mathbf{W}_{\text{q}}\cdot c^{\text{net}}(\mathbf{I}_{1}), \quad
    \mathbf{K}  =  \mathbf{W}_{\text{k}}\cdot c^{\text{net}}(\mathbf{I}_{2}),
\end{equation}
where symbol $\cdot$ stands for matrix product; $\mathbf{W}_{\text{q}}$ and $\mathbf{W}_{\text{k}}$ are learnable linear transformations. The \textit{Query} and \textit{Key} yield a cross-attention
\begin{equation}
    \mathbf{A}[i,j,k,l]=\sigma\left(\frac{\langle\mathbf{Q}[i,j,1:d], \mathbf{K}[k,l,1:d]\rangle}{\sqrt{d}} \right),
    \label{eq:attention}
\end{equation}
where $d$ is the feature length of $\mathbf{Q}[i,j,:]$ and $\mathbf{K}[k,l,:]$. The normalization function $\sigma$ for the an attention mechanism generally takes the form of \textit{softmax} mapping in previous works. However, in our work the cross-attention $\mathbf{A}$ serves to gate false positive correlation values between matching features, therefore a \textit{sigmoid} function performs better (see Sec.~\ref{sec:ablation} Ablation Study). 

We define the \textit{Value} matrix as the identity transformation of the traditional correlation volume $\mathbf{C}$ constructed by Eq.~\ref{eq:CostVolumeLevel1}. The element-wise production between $\mathbf{A}$ and $\mathbf{C}$ obtains a context aware correlation volume
\begin{equation}
    \mathbf{M} = \mathbf{A}\odot \mathbf{C}.
\end{equation}

\noindent\textbf{Lifting Operation} The above gating scheme can effectively screen false positives, leveraging the pairwise relevancy inferred from semantic context. Yet on the other hand, a true pair of correspondences may still under-estimate their correlation, if the non-normalized matching features $g(\mathbf{I}_{1})[i,j]$ and $g(\mathbf{I}_{2})[i^{\prime},j^{\prime}]$ have small magnitudes. In this situation, their inner product may cause $\mathbf{M}[i,j, i^{\prime},j^{\prime}]$ to be small relatively to correlation values estimated between false positives, since the gating strength of $\mathbf{A}$ is bounded by 1. As a consequence, the flow decoder may still mistake a false correspondence with higher correlation as the true correspondence. 

To lift the small correlation values between potential false negatives, we superimpose a scalar-weighted cross-frame correlation volume constructed by semantic context $\mathbf{S}$ to the gated correlation volume $\mathbf{M}$:
\begin{equation}
    \mathbf{S}[i,j,k,l] = \frac{\langle c^{\text{net}}(\mathbf{I}_{1})[i,j,1:t], c^{\text{net}}(\mathbf{I}_{2})[k,l,1:t]\rangle}{\sqrt{t}},
    \label{eq:contextCost}
\end{equation}
\begin{equation}
\mathbf{V} = \mathbf{M} + \lambda\times\mathbf{S},
\label{eq:CMCV}
\end{equation}
where the scalar $\lambda$ is a learned parameter initialised to zero. We observed that the trained $\lambda$ is always of the order of ${10^{-2}}$, at different training stages. Thus the semantic context correlation volume $\mathbf{S}$ only impacts $\mathbf{V}$ if $c^{\text{net}}(\mathbf{I}_{1})[i,j,:]$ and $c^{\text{net}}(\mathbf{I}_{2})[k,l,:]$ exhibit strong similarity. Moreover, we find that $\mathbf{S}$ enables the context attention to focus better (see our ablation study evidence Sec.~\ref{sec:ablation}). 

\noindent\textbf{Plug-and-Play Integration} Eq.~\ref{eq:CMCV} defines the proposed Context Guided Correlation Volume (CGCV), which forms the first level of the correlation pyramid to sample the correlation features. It only needs to be computed once in the forward pass. Each time the flow field is refined, the new correlation features are retrieved from the correlation volume without recomputing feature inner products. 

The proposed CGCV can be integrated with any RAFT-based optical flow neural networks, by simply replacing the traditional correlation volume, while remaining the rest architecture unaltered. In this work, we integrate CGCV with GMA. Fig.~\ref{fig:CVM} depicts the whole computation diagram of our flow computation model.

\section {Experiments}
\label{sec:experiments}

\begin{table*}[!h]\centering
\setlength{\tabcolsep}{1.mm}{
\begin{tabular}{@{}llccccccccccccccccc@{}}\toprule
{\multirow{2}{*}{Training }}  & {\multirow{2}{*}{Method}}& \multicolumn{2}{c}{Sintel (train)} & \phantom{a}& \multicolumn{2}{c}{KITTI-15 (train)} & \phantom{a} & \multicolumn{2}{c}{Sintel (test)} & \phantom{a} & \multicolumn{2}{c}{KITTI-15 (test)}& \phantom{a} &{\multirow{2}{*}{Params}}\\
\cmidrule{3-4} \cmidrule{6-7} \cmidrule{9-10} \cmidrule{12-13}
{Data} & {} & Clean & Final && AEPE & F1-all ($\%$) && Clean & Final && \multicolumn{2}{c}{F1-all ($\%$)} \\ \midrule
{\multirow{9}{*}{C+T}} & {DPCTF-F \cite{deng2021detail}} & {2.04} & {5.37}  && {8.42} & {19.93} && -&-&&\multicolumn{2}{c}{-}&&{38.6M}  \\ 
{} & {DEQ-RAFT-H \cite{bai2022deep}} & {1.41} & {2.75}  && \textbf{4.38} & \textbf{14.9} && -&-&&\multicolumn{2}{c}{-}&& {12.8M}  \\ 
{} & {RAFT+OCTC \cite{jeong2022imposing}} & {1.31} & \textit{2.67}  && {4.72} & {16.3} && -&-&&\multicolumn{2}{c}{-}&& \textbf{5.3M}  \\ 
{} & {GMA \cite{Jiang2021GMA}} & {1.30} & {2.74}  && {4.69} & {17.1}&& -&-&&\multicolumn{2}{c}{-} && {5.9M}  \\ 
{} & {AGFlow \cite{AGFlow}} & {1.31} & {2.69} && {4.82} & {17.0}&& -&-&&\multicolumn{2}{c}{-} && \textit{5.6M}\\
{} & {CRAFT \cite{sui2022craft}} & {1.27} & {2.79} && {4.88} & {17.5}&& -&-&&\multicolumn{2}{c}{-} && {6.3M}\\
{} & {FlowFormer-S \cite{flowformer}} & \textit{1.20} & \textbf{2.64} && {4.57} & {16.6}&& -&-&&\multicolumn{2}{c}{-} && {6.2M} \\
{} & {KPA-Flow \cite{luo2022learning}} & {1.28} & {2.68} && \textit{4.46} & \textit{15.9}&& -&-&&\multicolumn{2}{c}{-} && {5.8M}\\
{} & {CGCV (ours)} & \textbf{1.15} & {2.70} && {4.61} & {16.6}&& -&-&&\multicolumn{2}{c}{-} && {5.9M}\\ \midrule
{} & {SABMFL\cite{zhang2021self}} &  - & -  && - & - && {4.48} & {4.77} && \multicolumn{2}{c}{7.68}&&- \\
{} & {DPCTF-F\cite{deng2021detail}} &  ({0.81}) & ({1.16})  && ({1.34}) & ({7.3}) && {3.54} & {4.47} && \multicolumn{2}{c}{7.22}&&38.6M \\
{} & {DEQ-RAFT-H \cite{bai2022deep}} & {(0.70)} & {({1.21})}  && {({0.61})} & {({1.4})} && {1.82} & {3.23} && \multicolumn{2}{c}{4.98}&&12.8M   \\
{} & {RAFT+OCTC \cite{jeong2022imposing}} & {(0.73)} & {({1.23})}  && {({0.67})} & {({1.7})} && {1.82} & {3.09} && \multicolumn{2}{c}{\textit{4.72}}&&\textit{5.3M}   \\
{\multirow{3}{*}{C+T+}} & {GMFlow \cite{xu2022gmflow}} & {-} & {{-}}  && {{-}} & {{-}} && {1.74} & {2.90} && \multicolumn{2}{c}{9.32}&&\textbf{4.7M}   \\
{} & {GMFlowNet \cite{zhao2022global}} & {(\textbf{0.59})} & {(\textbf{0.91})}  && {({0.64})} & {({1.5})} && \textit{1.39} & {2.65} && \multicolumn{2}{c}{{4.79}}&&9.3M   \\
{S+K+H} & {GMA \cite{Jiang2021GMA}} & {(0.62)} & {({1.06})}  && {({0.57})} & {(\textit{1.2})} && \textit{1.39}* & {2.47*} && \multicolumn{2}{c}{5.15}&&5.9M   \\ 
{} & {AGFlow \cite{AGFlow}} & {(0.65)} & {(1.07)} && {(0.58)} & (\textit{1.2}) && {{1.43}*} & {2.47*} && \multicolumn{2}{c}{{4.89}}&&{5.6M}\\
{} & {CRAFT \cite{sui2022craft}} & {(\textit{0.60})} & ({1.06})  && {(0.58)} & {({1.3})} && {1.45*} & {\textit{2.42}*} && \multicolumn{2}{c}{{4.79}}&&6.3M\\ 
{} & {KPA-Flow \cite{luo2022learning}} & {(\textit{0.60})} & (\textit{1.02})  && {(\textbf{0.52})} & {(\textbf{1.1})} && \textbf{1.35}* & {\textbf{2.36}*} && \multicolumn{2}{c}{\textbf{4.60}}&&5.8M\\ 
{} & {CGCV (ours)} &  ({0.61}) & ({1.06})  && (\textit{0.56}) & ({1.3}) && \textbf{1.35}* & {2.43*} && \multicolumn{2}{c}{4.96}&&5.9M \\
\bottomrule
\end{tabular}}
\vspace{1ex}
\caption{Seven sets of performance comparisons about CGCV models with comparable models, all fine-tuned at the same stage. Results of competing methods are as reported in their original publications. Smaller value means better performance. Bold numbers indicate the highest accuracy; Italic numbers indicate second-best accuracy. Parentheses indicate the ground truth is released to public. Symbol * means using ``warm start'' \cite{teed2020raft}. }
\label{tab:CTSKH}
\end{table*}

\begin{table}[!h]\centering
\setlength{\tabcolsep}{1.mm}{
\begin{tabular}{@{}lcccccc@{}}\toprule
{\multirow{2}{*}{Method}}&  {Sintel (test)} & \phantom{a}& {Sintel (test) } & \phantom{a} & {KITTI-15} \\
\cmidrule{2-2} \cmidrule{4-4} \cmidrule{6-6}
{}&{Clean} & \phantom{a}& {Final} & \phantom{a} & {(test)} \\
\midrule
{GMA\cite{Jiang2021GMA}}&{12}&&{11}&&{71}\\
{AGFlow\cite{AGFlow}}&{19}&&{10}&&{43}\\
{CRAFT\cite{sui2022craft}}&{21}&&{7}&&{33}\\
{CGCV (Ours)}&{9}&&{8}&&{48}\\

\bottomrule
\end{tabular}}
\vspace{1ex}
\caption{The MPI-Sintel and KITTI Leader Board ranking of the proposed CGCV and its comparable state of the art methods at the time of writing. Among competing methods, CGCV ranks the highest on the Clean pass, the second-highest on the Final pass, and the third on KITTI-15.}
\label{tab:Ranking}
\end{table}

\begin{figure*}[h]
\centering
\begin{tabular}{cccc}
 \includegraphics[width=0.225\linewidth]{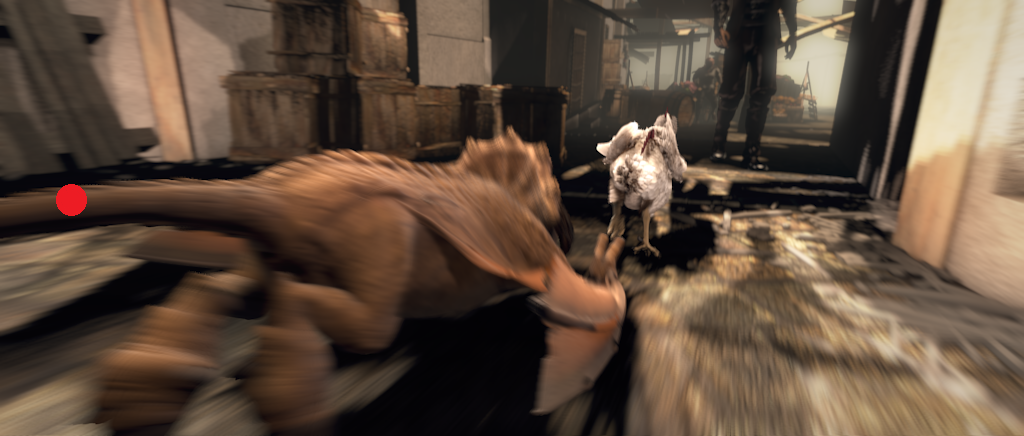} &
 \includegraphics[width=0.225\linewidth]{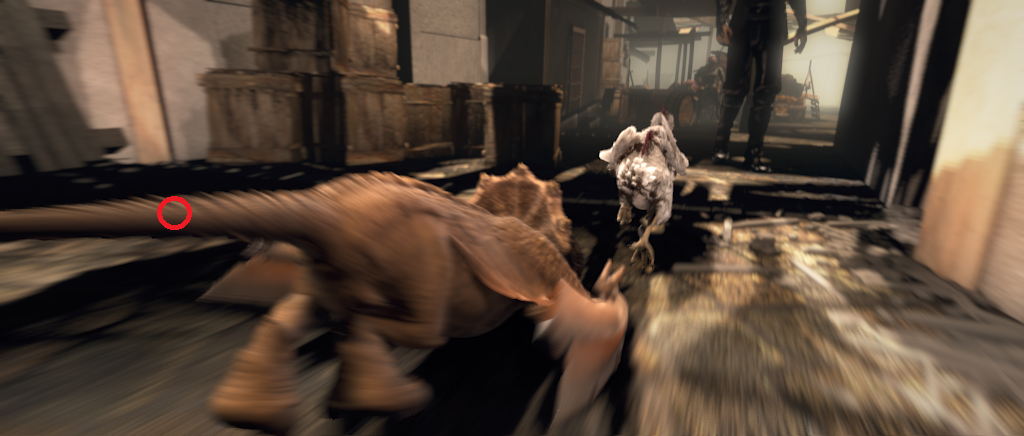} &
 {\includegraphics[width=0.225\linewidth]{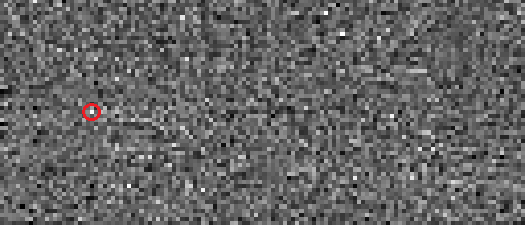}} &
 \includegraphics[width=0.225\linewidth]{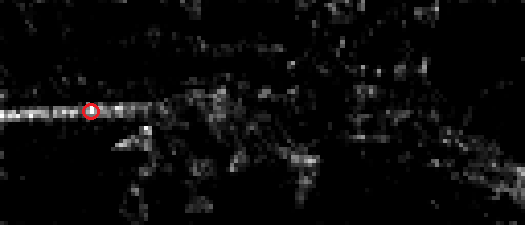} \\
 \small{(a) Reference} & \small{ (b) Target}  & \small{(c) Matching Correlation} & \small{(d) Context Attention}\\
 \includegraphics[width=0.225\linewidth]{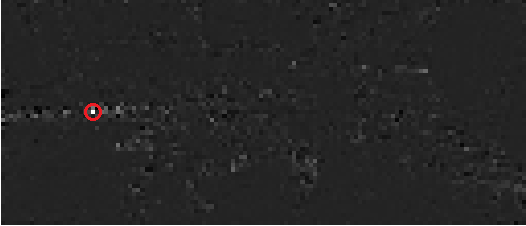}&
 \includegraphics[width=0.225\linewidth]{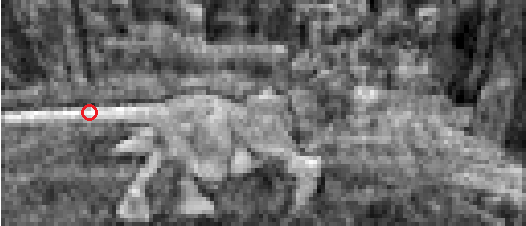} &
 \includegraphics[width=0.225\linewidth]{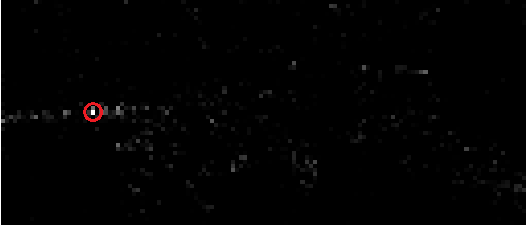} & 
 \includegraphics[width=0.225\linewidth]{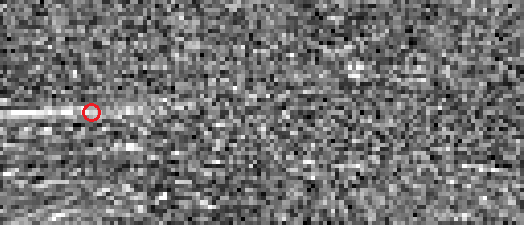}\\
  \small{(e) Context Gated Correlation} & \small{(f) Context Correlation}& \small{(g) CGCV} & \small{ (h) GMA Correlation}\\
  \multicolumn{2}{r}{\includegraphics[width=0.475\linewidth]{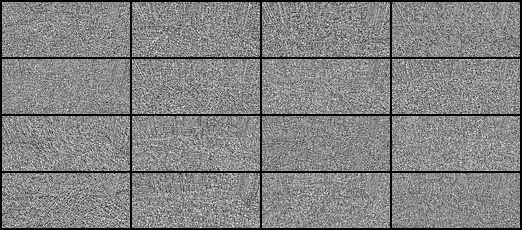}}
  &
  \multicolumn{2}{l}{\includegraphics[width=0.475\linewidth]{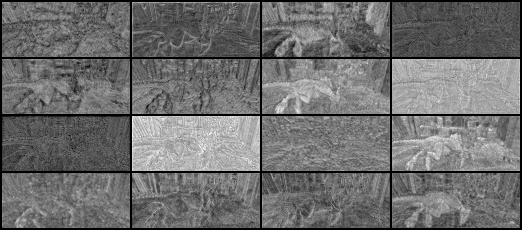}}
  \\
  \multicolumn{2}{c}\small{(i) Matching Feature Channels of Reference Frame} & 
  \multicolumn{2}{c}\small{(j) Context Feature Channels of Reference Frame}

\end{tabular}
\caption{A visualization of how our regularization method improves the traditional correlation volume, taking an example \textit{query} point (indicated as the red dot on the reference frame) from the tail of the dragon in test image \textit{Market 4} of Sintel (test) Final. The visualization is conducted using the CTSKH-tuned CGCV. Red circles in sub-figures indicate the true correspondence positions for the query point. Sub-figure (g) shows the consistency between the maximum CGCV value position and the true correspondence position. In contrast, GMA correlation plane exhibits large values at many false correspondence positions. For clarity, the brightness of each involved map is normalized over the lattice. Better viewed digitally. }
 \label{fig:maps_1}
\end{figure*}

\begin{figure*}[h]
\centering
\begin{tabular}{cccc}
 \includegraphics[width=0.225\linewidth]{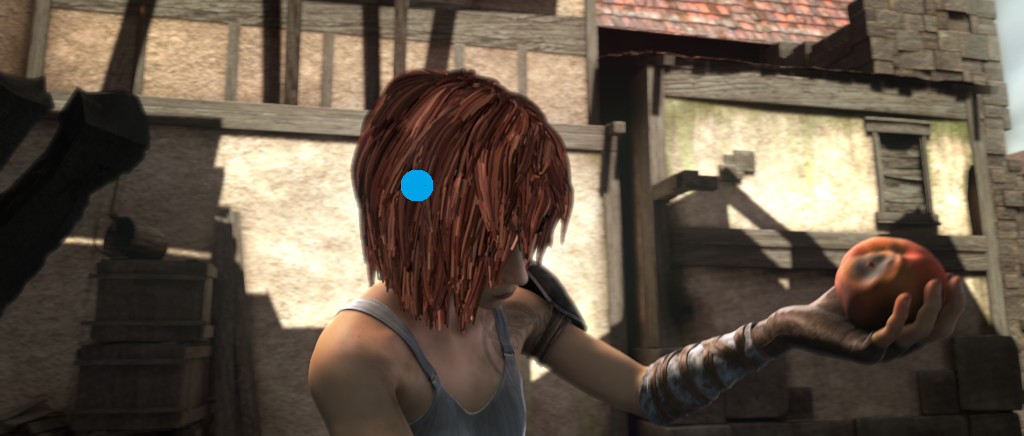} &
 \includegraphics[width=0.225\linewidth]{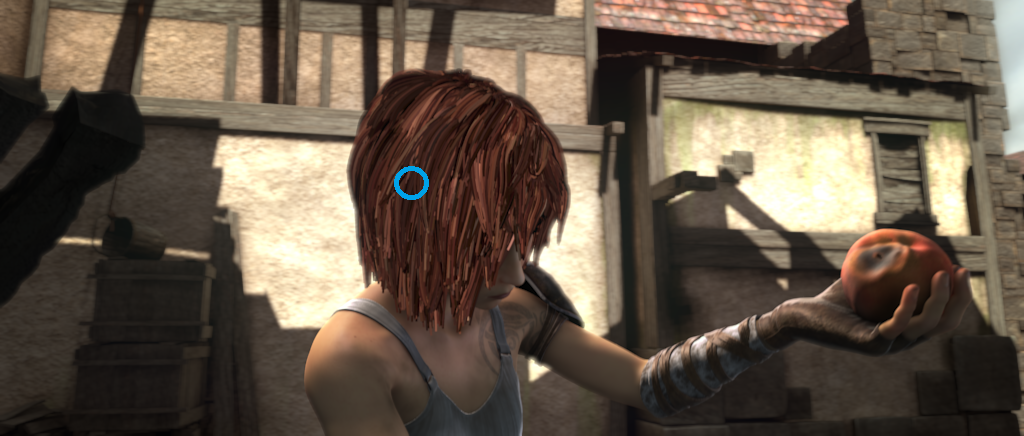} &
 \includegraphics[width=0.225\linewidth]{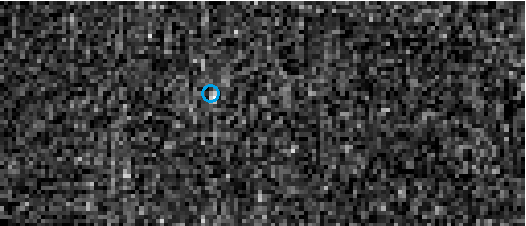} &
 \includegraphics[width=0.225\linewidth]{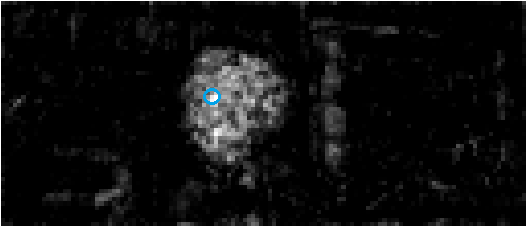} \\
 \small{(a) Reference} & \small{(b) Target} & \small{(c) Matching Correlation} & \small{(d) Context Attention}\\
 \includegraphics[width=0.225\linewidth]{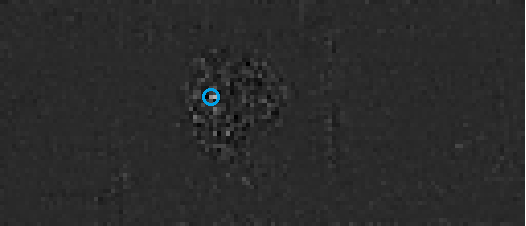}&
 \includegraphics[width=0.225\linewidth]{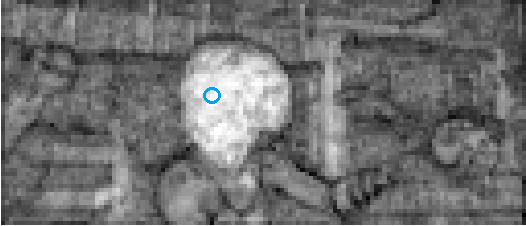} &
 \includegraphics[width=0.225\linewidth]{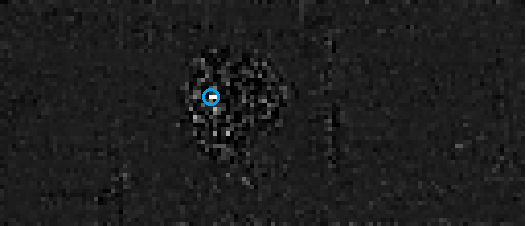} &
 {\includegraphics[width=0.225\linewidth]{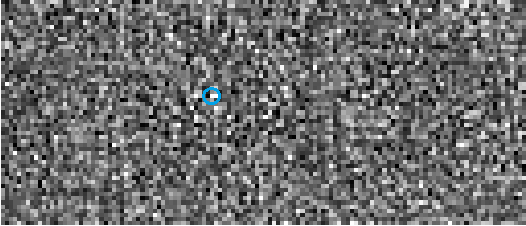}}\\
 \small{(e) Context Gated Correlation} & \small{(f) Context Correlation}& \small{(g) CGCV} & \small{ (h) GMA Correlation}\\
\end{tabular}
\caption{An example to visualize the development process of the porposed CGCV correlation volume. The query point is taken from the red hair in train image \textit{Alley 1} of Sintel (train) Final (indicated as the blue dot on the reference frame). The visualization is conducted using the CTSKH-tuned CGCV. Blue circles in sub-figures indicate the true correspondence positions for the query point. Sub-figure (g) shows the consistency between the maximum CGCV value position and the true correspondence position. In contrast, GMA correlation plane exhibits large values at many false correspondence positions. For clarity, the brightness of each involved map is normalized over the lattice. Better viewed digitally.}
 \label{fig:maps_2}
\end{figure*}

\begin{table*}[!h]
\centering
\begin{tabular}{@{}lccccccc@{}}
\toprule
\multirow{3}{*}{Method}&Chairs& \phantom{a} &\multicolumn{2}{c}{Things}& \phantom{a} &\multicolumn{2}{c}{Sintel} \\
\cmidrule{4-5} \cmidrule{7-8}
&&&Clean&Final&&Clean&Final \\
&(val)&&(test)&(test)&&(train)&(train) \\
\midrule
{CGCV}&\textbf{0.69}&&\textit{2.60}&\textit{2.39}&&\textbf{1.15}&\textbf{2.70} \\
{removing context correlation volume}&\textit{0.70}&&\textbf{2.47}&\textbf{2.35}&&\textit{1.20}&\textit{2.71}\\
{replacing sigmoid attention by softmax}&0.73&&2.95&2.69&&1.32&3.09 \\
{removing context attention}&0.81&&3.02&2.61&&1.30&2.74 \\
\bottomrule
\end{tabular}
\vspace{1ex}
\caption{Ablation study on the proposed CGCV module, the main components of which are: a Context Cross-Frame Attention Module with a Sigmoid function, a Context Cross-Frame Correlation Volume. Bold numbers indicate the top accuracy. Slanted numbers indicate the second-best accuracy.}
\label{tab:ablation}
\end{table*}

\subsection{Settings}
The hardware platform for our experiments is two Nvidia RTX 2080Ti graphics cards, and the software environment is PyTorch. To evaluate the effectiveness and efficiency of the proposed Context Guided Correlation Volume (CGCV), we adopt GMA as our baseline framework and replace its traditional correlation volume with CGCV. For fair evaluation, we set the matching radius $r$ to 4 and the context feature length $c$ (Eq.~\ref{eq:contextCost}) to 256, the length of the inp and net features is set to 128,  as same as GMA. The length of the query and key features $d$ (Eq.~\ref{eq:attention}) is also set to 128.

Our training and testing conform to the commonly used 4-phases procedure in recent literature of flow computation neural models \cite{teed2020raft,Jiang2021GMA,AGFlow,sui2022craft,flowformer}. At the first phase, our model is trained on FlyingChairs for 120k iterations with a batch size of 8. The second phase continues the training on FlyingThings3D for 120k iterations with a batch size of 6. The third phase fine-tunes the pre-trained model on the comprehensive dataset composed of FlyingThings3D, Sintel (train), KITTI (train) and HD1K for 120k iterations with a batch size of 6. The forth phase further fine-tunes the phase-3 model on KITTI (train) for 50K iterations with a batch size of 6. Our learning rate schedule complies with GMA: the maximum learning rate is set to $0.25\times 10^{-3}$ for the first phase, and $0.125\times 10^{-3}$ for the other phases. For clarity, we call the models learned at phase 2-4 the \textit{CT-trained}, \textit{CTSKH-tuned} and \textit{KITTI-tuned} models, if specification is necessary, hereafter in this paper. 

The benchmark datasets and evaluation metrics in our experiments follow the literature convention. The CT-trained model is evaluated on the test datasets officially split from Sintel (train) and KITTI-15 (train), to assess the generalization ability of our model. The CTSKH-tuned model is tested on Sintel (train) and Sintel (test). The KITTI-tuned model is tested on KITTI-15 (train) and KITTI-15 (test).

\subsection{Quantitative Analysis}
Table \ref{tab:CTSKH} lists flow computation neural methods published in recent couple of years. Compared to our method, DPCTF-F \cite{deng2021detail}, DEQ-RAFT-H \cite{bai2022deep} and GMFlowNet \cite{zhao2022global} have significantly more parameters; KPA-Flow \cite{luo2022learning} requires notably more FLOPs; GMFlow \cite{xu2022gmflow} and GMFlowNet \cite{zhao2022global} formulate flow computation without refinement. Although these methods are not directly comparable to ours, they provide a frame of reference for top performance to the readers.   

\noindent\textbf{Evaluation on Leader Boards.} Our model is evaluated by the MPI-Sintel \cite{Sintel} and KITTI-15 \cite{KITTI} Leader Boards by their test data and metrics. Among the 11 competing methods, ours ranks the 1st on Sintel (test) Clean, the 3rd on Sintel (test) Final, and the 5th on KITTI-15. 

\noindent\textbf{Evaluation of Model Generalization} At the C+T training stage, with respect to other 8 top-performing methods, ours ranks the 1st place on Sintel (train) Clean, the 5th on Sintel (train) Final, and the 4th place on KITTI. This validates that our model generalizes well to unseen datasets.

\noindent\textbf{Evaluation of Fine-Tuned Models.} At the CTSKH fine-tuning stage, among the 9 compared methods, ours ranks the 2nd on KITTI-15 (train) in AEPE, the 3rd on Sintel (train) Final, and the 3rd on Sintel (train) Clean. Its accuracy score is very close to methods of similar complexity.


\noindent\textbf{Comparative Analysis} In the literature, GMA, AGFlow \cite{AGFlow}, CRAFT \cite{sui2022craft} adopt the same paradigm and have similar complexities \footnote{Although KPA-Flow \cite{luo2022learning} is also based on GMA and have similar parameter counts, it runs substantially slower than the proposed model. For example on our machine, it takes KPA-Flow 0.30 seconds to compute the flow of a pair of KITTI images, whereas it take CGCV 0.20 seconds in identical settings. Hence here we do not compare to KPA-Flow.}. Table \ref{tab:Ranking} presents the MPI-Sintel and KITTI Leader Boards Ranking of CGCV and these comparable methods. In this comparison, CGCV achieves the best score on Sintel (test) Clean, the second-best score On Sintel (test) Final (slightly lower than CRAFT). We now analyze our method with respect to these methods one by one. CGCV notably improves its baseline model GMA in all evaluations. 

\textbf{CGCV vs GMA.} Our goal is to improve the traditional correlation volume in the presence of distracting factors. As our model replaces the correlation volume of GMA by the proposed one, a comparison with GMA would reveal the effectiveness of our correlation volume design. CGCV (5.91~M) has a similar size to GMA (5.88~M). At the CT-training stage, CGCV has higher accuracy than GMA in all 4 sets of comparisons. Especially on Sintel (train) Clean, CGCV reduces the AEPE of GMA by $11.53\% (1.30\rightarrow{1.15})$ times. After CTSKH fine-tuning, CGCV and GMA show similar performance on both Sintel and KITTI. However, in Sintel and KITTI Leader Boards evaluation, CGCV shows clear superiority: it increases GMA by 23 places in KITTI Ranking. 

\textbf{CGCV vs AGFlow.} Both AGFlow and our method aims at enhancing motion estimation by context information. While our method turns to semantic context cross-frame attention, AGFlow is based on graph reasoning. CGCV is only 0.3~M larger than AGFlow in model parameter scale, but CGCV scores better in 9 out of the total 11 sets of comparisons. Noticeably on Sintel (train) Clean, the CT-trained CGCV model has $12.21\%$ $({1.31}\rightarrow{1.15})$ times lower AEPE; On Sintel (test) Clean, CGCV has $5.59\%$ $ ({1.43}\rightarrow{1.35})$ times lower AEPE. CGCV is slightly less accurate than AGFlow by $1.43\%$ $({4.89}\rightarrow{4.96})$ times on KITTI-15 (test). These experiments verify the competitiveness of our context cross-frame attention to graph reasoning, in leveraging context to enhance motion tokens.

\textbf{CGCV vs CRAFT.} Both CRAFT and our method adopt the framework of GMA. CRAFT applies Transformer to the matching features by dynamically aggregating mode attention, whereas we apply context cross-frame attention. CGCV is 0.4~M smaller in size. Numerically, CGCV shows superiority on 6 out of total 11 sets of comparisons. The most notable improvements occur at CT-training stage on 
Sintel (train) Clean and CTSKH fine-tuning stage on Sintel (test) Clean, where the errors of CGCV are $9.45\%$ $({1.27}\rightarrow{1.15})$ and $6.90\%$ $({1.45}\rightarrow{1.35})$ times better. Moreover, on KITTI-15 (train), the AEPE and F1-all scores of CGCV are $5.53\%$ $({4.88}\rightarrow{4.61})$ and $5.14\%$ $({17.5}\rightarrow{16.6})$ times better. This indicates that CT-trained CGCV generalizes better to unseen inputs. The accuracy of CGCV is slightly $3.55\%$ $({4.79}\rightarrow{4.96})$ times weaker on KITTI-15 (test). The comparative analysis shows that our context regulation strategy is competitive with CRAFT's Transformer-based strategy in flow computation. 

\textbf{CGCV vs FlowFormer-S.} FlowFormer is a highly effective deep neural model in Transformer architecture with 18.2~M parameters. To fairly compare to GMA, a small version FlowFormer-S, which has 6.2~M parameters, is trained on C+T without fine-tuning and evaluated in \cite{flowformer}. Compared to FlowFormer-S, CGCV is 0.3~M smaller in size. CGCV exhibits superiority by $4.16\%$ $({1.20}\rightarrow{1.15})$ times smaller AEPE on Sintel (train) Clean and light inferiority by $2.27\%$ $({2.64}\rightarrow{2.70})$ times larger APAE on Sintel (train) Final. In these experiments, the convolutional network architecture of CGCV obtains comparable performance to the transformer architecture of FlowFormer-S.

\subsection{Qualitative Analysis}
\noindent\textbf{Visualization of Flow Fields.} Fig.~\ref{fig:GraphAbs} visualizes the flow fields estimated by CGCV and GMA on three example images from Sintel (test) Final. Image \textit{Mountain 2} contains a large textureless region undergoing fast camera motion. As a consequence, the flow estimation of GMA in this region suffers large error, but it can be substantially improved by regulating its correlation volume using context. On image \textit{Tiger}, GMA mis-estimates the motion in the shadowed ground region with large error, whereas CGCV computes it mostly correct. On image \textit{Market 3}, GMA blurs the motion boundary between the out-of-focus foreground object and the foggy background. In contrast, the flow estimated by CGCV in this region is sharp. These visual comparisons validate that regulating the traditional correlation volume by semantic context increases the flow computation robustness to image flatness, shadows, de-focus blur and atmospheric effects.

\noindent \textbf{Visualization of Feature Maps and Correlation Volumes.} Fig.~\ref{fig:maps_1} and Fig.~\ref{fig:maps_2} illustrate how our regularization approach improves the quality of the traditional correlation volume for flow computation sequentially, through two example query points (one on the dragon tail, one on the red hair) in image \textit{Market 4} from Sintel (test) Final and image \textit{Alley 1} from Sintel (train) Final. Here we take the CTSKH-tuned GMA and CGCV models to visualize the correlation maps. 

As demonstrated by Fig.~\ref{fig:maps_1}, in either GMA or our CGCV, the correlation volumes computed from the matching features are very noisy. A large number of pixels, which scatter over the target frame, exhibit high correlation to the query point. However, the context cross-frame attention map successfully figures out that only pixels on the dragon tail in the target image have high context correlation to the query point. Weighting the matching correlation by context attention effectively screens the false candidates for the true correspondence, with the highest correlation values locating on the dragon tail. It can be seen that the context feature correlation map has significantly superior quality to the matching correlation map. Although superimposing it to the context-weighted correlation map does not contribute much to the final correlation map, it benefits obtaining a sharp attention map, thereby forming the high quality final correlation volume.  

Similar correction effects can be observed in Fig.~\ref{fig:maps_2}. Our gating strategy effectively narrows down the high correlation regions from the whole image to the red hair area of the image, where the true correspondence to the query point locates. 


\subsection{Ablation Study}
\label{sec:ablation}
An ablation study is carried out using the CT-trained model. We gradually remove the main components that construct our CGCV module, until the correlation volume takes the traditional form as in GMA and RAFT, and evaluate the degenerated model step by step on benchmark datasets employed by GMA for ablation study \cite{Jiang2021GMA}. Accuracy is measured by AEPE.

We first remove the context cross-frame correlation volume from the whole correlation volume (i.e., degenerate $\beta$ in Eq.~\ref{eq:CMCV} to 0). This ablation increases the flow accuracy on Thing (test) Clean and Thing (test) Final by 0.13 and 0.04 pixels. However, when generalizing to Sintel, the ablation causes performance degradation. Based on this observation, we hypothesis that the context cross-correlation benefits the generalization ability of the model, and hence adopt it as the lifting term of CGCV. 

We further replace the Sigmoid function in the context cross-frame attention (Eq.~\ref{eq:attention}) by a softmax function. This drastically lowers the flow accuracy on all datasets. The AEPE on Things (test) Clean and Final deteriorates by 0.48 pixels and 0.34 pixels respectively. On Sintel (train) Clean, the AEPE retreats by 0.12 pixels. At this stage, accuracy measurements of the degenerated model in all tests are remarkably lower than CGCV.

Finally, we switch off the context cross-attention gating mechanism, as defined in Eq.\ref{eq:CostVolumeLevel1}. This  lowers the flow accuracy on Sintel (train) Clean by 0.1 pixels.

The ablation study verifies that all composing components of the new correlation volume are crucial to improve the quality of traditional correlation volume.

\section{Conclusion}
In this paper, we have presented a novel correlation volume for RAFT flow computation paradigm. We take advantage of the semantic context features to suppress the false positive matching in traditional cost volume via cross-attention, and lift the false negative matching via cross-correlation. The presented correlation volume is simple, light-weight, highly effective and ready to replace the cost volume in state of the art RAFT-style flow computation neural models. By incorporating our cost volume with the GMA-RAFT framework, we have shown that our cost volume substantially enhances the numerical and visual performance of GMA, with superiority to other GMA-based methods, by comprehensive experiments.

{\small
\bibliographystyle{cvm}
\bibliography{egbib}
}

\end{document}